\pdfoutput=1

\documentclass[11pt]{article}

\usepackage{acl2023}

\usepackage{times}
\usepackage{latexsym}
\usepackage{graphicx}
\usepackage[T1]{fontenc}

\usepackage[utf8]{inputenc}

\usepackage{microtype}

\usepackage{inconsolata}

\title{USA: Universal Sentiment Analysis Model \& Construction of Japanese Sentiment Text Classification and Part of Speech Dataset}

\author{
  Chengguang Gan\textsuperscript{1} \quad
  Qinghao Zhang\textsuperscript{2} \quad
  Tatsunori Mori\textsuperscript{1} \\
  \textsuperscript{1}Yokohama National University, Japan \\
  \texttt{\{gan-chengguan-pw@ynu.jp, tmori@ynu.ac.jp} \\ynu.jp
  \textsuperscript{2}Department of Information Convergence Engineering, \\
  Pusan National University, South Korea \\
  \texttt{zhangqinghao@pusan.ac.kr}
}

\begin{document}
\maketitle

\begin{abstract}
Sentiment analysis is a pivotal task in the domain of natural language processing. It encompasses both text-level sentiment polarity classification and word-level Part of Speech(POS) sentiment polarity determination. Such analysis challenges models to understand text holistically while also extracting nuanced information. With the rise of Large Language Models(LLMs), new avenues for sentiment analysis have opened. This paper proposes enhancing performance by leveraging the Mutual Reinforcement Effect(MRE) between individual words and the overall text. It delves into how word polarity influences the overarching sentiment of a passage. To support our research, we annotated four novel Sentiment Text Classification and Part of Speech(SCPOS) datasets, building upon existing sentiment classification datasets. Furthermore, we developed a Universal Sentiment Analysis(USA) model, with a 7-billion parameter size. Experimental results revealed that our model surpassed the performance of gpt-3.5-turbo across all four datasets, underscoring the significance of MRE in sentiment analysis. We already open sourced the USA-7B model at \href{https://huggingface.co/ganchengguang/USA-7B-instruction-incontext-learning}{Huggingface}\footnote{\url{https://huggingface.co/ganchengguang/USA-7B-instruction-incontext-learning}}.
\end{abstract}

\section{Introduction}

In sentiment analysis, many datasets and models predominantly focus on polarity classification of an entire text. Alternatively, they extract specific words from the text for sentiment polarity categorization. However, these methods often overlook the reciprocal relationship between individual words and the overall sentiment of the text. In other words, while specific words can influence the sentiment classification of an entire text, the overall sentiment can, conversely, shape the sentiment polarity of individual words within the text.

The concept of Mutual Reinforcement Effect(MRE) was initially introduced in the realms of sentence categorization and Named Entity Recognition(NER) tasks\cite{10.1007/978-3-031-35320-8_18}. Here, it was employed to simultaneously enhance the accuracy of both classifications by amalgamating the sentence classification and NER tasks. Drawing inspiration from this, our study aims to investigate the possible presence of MRE between the sentiment polarity categorization of texts and individual words.

\begin{figure}[t]
\centering
\includegraphics[width=219 pt]{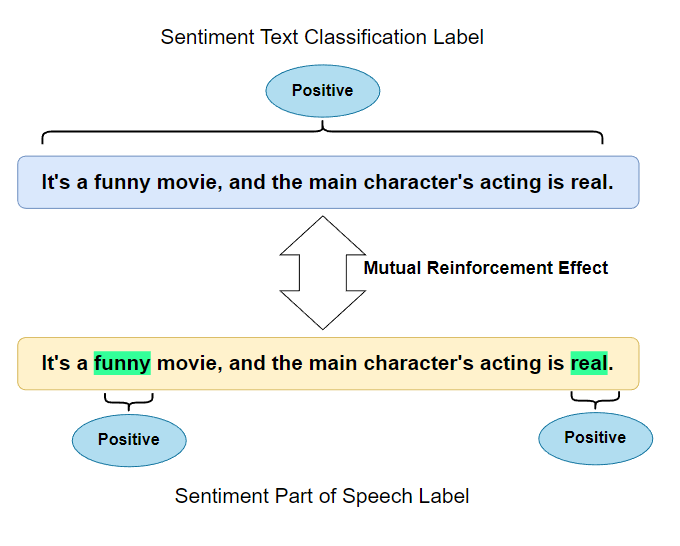}
\caption{Example of Mutual Reinforcement Effect in Sentiment Text Classification task and Part of Speech task.}
\label{figure1}
\end{figure}

As illustrated in Figure \ref{figure1}, the upper section represents the sentiment polarity categorization of the entire text, while the subsequent section delves into the sentiment polarity extraction and labeling of adjectives within the text. The examples in the figure demonstrate that if a text is classified as positive, then certain adjectives within the text are likely to lean towards a positive sentiment. Conversely, when a text contains a plethora of adjectives or nouns with positive polarity, the overarching sentiment of the text likely leans towards the positive. Recognizing this interplay provides an opportunity to merge these two tasks, enhancing the model's granular understanding of the text and thus improving the performance of both tasks simultaneously. This synergistic approach underpins our proposed Sentiment Text Classification and Part of Speech(SCPOS) dataset.

Moreover, there is a notable scarcity of datasets pertinent to sentiment analysis in the Japanese language. Currently, the MARC-ja dataset, part of the Japanese General Language Understanding Evaluation(JGLUE)\cite{kurihara2022jglue}, stands as the sole resource available for Japanese text sentiment classification. It is a binary classification dataset derived from user reviews on shopping websites. Remarkably, there is no dataset that offers sentiment polarity categorization at the word level for Japanese. The japanese itself presents unique characteristics distinct from other global languages. Not only is the Japanese script a combination of kanji, hiragana, and katakana, but its linguistic features—such as word transformations, sentence structures, and the nuanced use of honorifics—also set it apart. Given these intricacies, there's a compelling need to develop a dedicated dataset for Japanese sentiment analysis. This endeavor, therefore, addresses the existing deficiency in word-level sentiment polarity datasets for Japanese and also pioneers a fresh research trajectory in the SCPOS mixture task.

Otherwise, the advent of Large Language Models(LLMs) has heralded new avenues in sentiment analysis. When GPT-3\cite{brown2020language} was launched, it exhibited a remarkable zero-shot capability alongside an innate ability to predict subsequent sentences based on provided input samples. This feature is often referred to as "In-context learning". Such a capability is particularly advantageous for languages like Japanese that have limited datasets. Remarkably, this method allows for the benefits of fine-tuning on expansive datasets to be achieved with minimal manually labeled data. Furthermore, the introduction of ChatGPT\cite{ouyang2022training} showcased its robust generalization capabilities. LLMs possess extensive generalized knowledge and a depth of text comprehension that is challenging for smaller models to attain. This raises the intriguing question: Can the unique attributes of LLMs be leveraged to cultivate a generalized sentiment analysis model capable of handling both text sentiment analysis and Part-of-Speech(POS) sentiment polarity classification? Enhancements in word-level sentiment categorization and text-level sentiment classification could provide LLMs with heightened granularity and accuracy in discerning textual sentiments, thereby boosting their overall performance.

The remainder of this paper delineates the related work(§\ref{related work}), introduces the construction of the SCPOS dataset(§\ref{SCPOS}), explains the training of the USA model(§\ref{USA model}), presents an evaluation of both the baseline and USA models on the SCPOS dataset(§\ref{evaluation}), and analysis of MRE(§\ref{MRE}).

\section{Related Work}\label{related work}

\subsection{Initial studies in sentiment analysis}

Sentiment classification emerged as a specialized subset of text classification tasks\cite{pang2002thumbs}, with the primary objective of discerning the overall sentiment of a text as either negative or positive. In above study, employed machine learning methods such as naïve bayes, maximum entropy, and Support Vector Machines(SVM) to categorize the sentiment of texts. Notably, the research first time introduced movie reviews as a dataset for sentiment classification. Over the following decade, a plethora of datasets were developed for sentiment classification\cite{medhat2014sentiment}. These encompassed a range of sources including News Articles\cite{BAI2011732}, Hotel Reviews\cite{wu2011two}, Restaurant Reviews\cite{ROBALDO2013454}, Tweets\cite{go2009twitter}\cite{agarwal2011sentiment}, Blogs\cite{yu2013impact}, and more. As a result of these expanded datasets, sentiment classification algorithms have undergone significant evolution. They progressed from initial rule-based systems to statistical models, and further to Hidden Markov Models(HMM) and Conditional Random Fields(CRF). Such advancements have robustly shaped the foundation for future sentiment analysis endeavors.

\subsection{Deep Learning for Sentiment Analysis}

\begin{table*}[!t]
\centering
\begin{tabular}{lllll}
\hline
  & \textbf{SRW} & \textbf{NVA} & \textbf{N} & \textbf{VA} \\
\hline
 Label& positive, Xnegative, neutral, & positive, neutral, & positive, neutral, & positive, negative \\
 & Xpositive, negative& negative & negative &\\
 Count& 5 & 3 & 3 & 2 \\
 Total& 2000 & 187528 & 187528 & 187528 \\
\hline
\end{tabular}
\caption{\label{table1SCPOS data}
Statistical data of SCPOS. \textbf{NVA} dataset represents a dataset consisting of nouns, verbs and adjectives. \textbf{N} then represents consisting of nouns only. \textbf{VA} represents consisting of verbs and adjectives.}
\end{table*}

Over the past decade, there has been a significant surge in the use of neural networks in AI. This has ushered sentiment categorization into the realm of deep learning\cite{zhang2018deep}. Initially, researchers employed word embeddings and rudimentary neural network models for sentiment analysis. Subsequently, the advent of transformer models\cite{vaswani2017attention} marked the transition into the era of Pre-trained Language Models(PLMs), with BERT\cite{devlin2018bert} and T5\cite{raffel2020exploring} models enhancing the accuracy of text sentiment categorization considerably. Presently, we are in the era of Large Language Models(LLMs) with the introduction of models like GPT-3\cite{brown2020language}, ChatGPT\cite{ouyang2022training}, and GPT-4\cite{openai2023gpt4}, which claim to address a myriad of NLP challenges. However, when it comes to sentiment categorization accuracy, these LLMs often lag behind fine-tuned smaller models. We aspire for LLMs to leverage their extensive pre-training knowledge and comprehension when training a universal sentiment analysis model. This is crucial in ensuring high sentiment classification accuracy, especially in 0/1-shot scenarios.

\cite{li2023unisa} introduced UniSA, a unified generative framework to integrate sentiment analysis subtasks, addressing challenges like modality alignment, varied input/output formats, and dataset bias. They also curated SAEval, a benchmark that consolidates various sentiment subtask datasets. Their results highlighted UniSA's competitive performance across all subtasks in sentiment analysis. In contrast to previous research, this study emphasizes the relationship between individual words and the overall text. Specifically, it trains LLMs for sentiment analysis using a multi-task approach, distinguishing it from earlier studies.

\subsection{Resource of Japanese Sentiment Analysis}

The field of sentiment analysis in Japanese currently faces a significant dearth of resources. Existing datasets, such as one that categorizes shopping site reviews, offer only binary classifications\cite{kurihara2022jglue}. Remarkably, there is no specialized dataset available specifically for word-level sentiment polarity classification in Japanese. However, two sentiment polarity dictionaries do exist, focusing on common Japanese nouns and verb \& adjectives\cite{2005203}\cite{Adjectives}. Despite these dictionaries, word-level polarity classifications have not been applied to broader textual contexts. Thus, there is an evident and pressing need for a comprehensive dataset tailored for sentiment analysis in this language. Moreover, the endeavor to train generalized LLMs specifically for sentiment analysis remains uncharted territory. Previous efforts have largely centered around training universal Named Entity Recognition(NER) LLMs using open-domain NER datasets\cite{zhou2023universalner}. The outcomes of such 0-shot have proven to be commendable.

\begin{figure}[!h]
\centering
\includegraphics[width=219 pt]{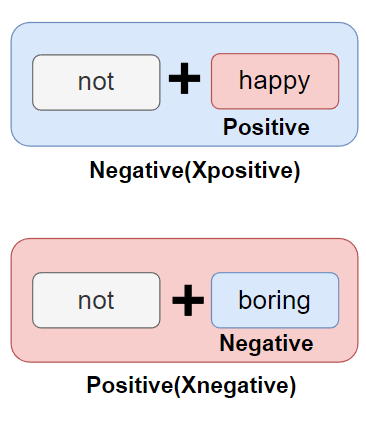}
\caption{Example of Xpositive and Xnegative label.}
\label{figure2}
\end{figure}

\begin{figure*}[!t]
\centering
\includegraphics[width=438 pt]{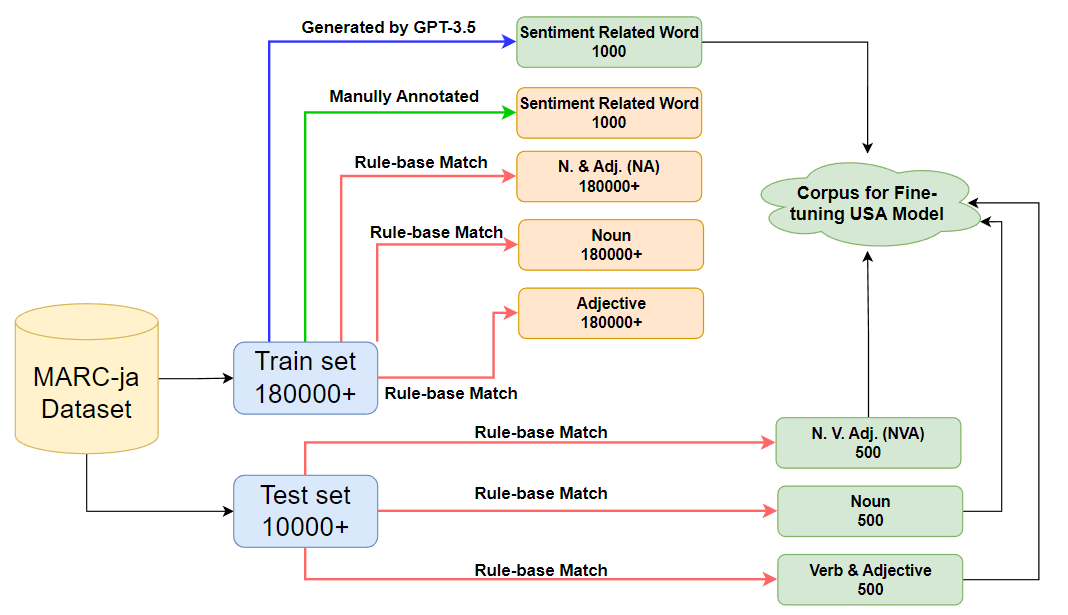}
\caption{Overview of SCPOS dataset construction process. Different colored lines correspond to different methods of construction. The number below the sub-dataset name is the dataset sample size.}
\label{figure3}
\end{figure*}

\section{Sentiment Text Classification and Part of Speech Dataset}\label{SCPOS}

In this chapter, we present the methodology employed in the creation of the SCPOS dataset. This elucidation facilitates a clearer comprehension when one proceeds to the training of the USA model. This chapter is organized into two subsections: the first addresses manual construction of the SCPOS datasets, while the second discusses its rule-based construction. All four SCPOS datasets are annotated based on the MARC-ja dataset from JGLUE. Table \ref{table1SCPOS data} provides statistical data regarding the SCPOS dataset, and Figure \ref{figure3} offers a visual representation of its composition and labeling process, encompassing both test set and train corpus.

\subsection{SCPOS dataset of Manually Annotated and LLM generated}

In this section, we delineate the process of constructing the Sentiment Related Word (SRW) dataset and outline the method for utilizing manually annotated data to train the LLM, enabling it to autonomously annotate the dataset. Within the manually annotated dataset, emphasis is placed on words that significantly influence the overarching sentiment polarity of the text. Consequently, individual words are not labeled for their specific sentiment polarity. This approach was adopted as adjectives play a predominant role in determining the text's overall sentiment. As a result, all adjectives in the SRW dataset were annotated. Furthermore, nouns and verbs associated with either a negative or positive sentiment were annotated, while those deemed neutral were excluded from annotation.

\begin{figure*}[!t]
\centering
\includegraphics[width=438 pt]{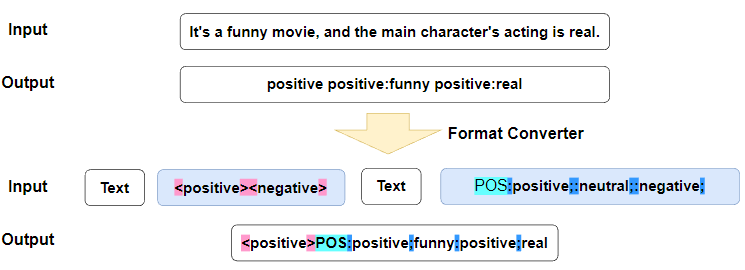}
\caption{Example of datasets input and ouput format.}
\label{figure4}
\end{figure*}

A conventional POS sentiment polarity categorization typically encompasses 2-3 labels, such as Positive, Neutral, and Negative. To this taxonomy, we introduce two additional labels: Xnegative and Xpositive. As illustrated in Figure \ref{figure2}, the sentiment conveyed by a positive adjective, like "happy," is inverted when preceded by a negative word, turning the otherwise positive adjective into a negative context. Conversely, a negative adjective, such as "boring," when preceded by a negative word, can be transformed to convey a positive sentiment. Recognizing these nuances is crucial in accurate sentiment polarity determination. This is particularly salient in Japanese where the linguistic structures and words indicating negation are both diverse and markedly distinct from many other languages. Such complexities may lead the model to misinterpret the sentiment polarity of words or entire texts. Consequently, by introducing these two labels, we aim to enhance the model's proficiency in understanding the lexical shifts of adjectives in the presence of negation, ensuring more accurate sentiment polarity assessments. It's also worth noting that, with the addition of negative words or phrases, multiple word might be required for a single label, extending to two words or even a phrase.

Following the aforementioned annotation rules, we obtained 1,000 samples, constituting a high-quality, manually annotated dataset. We utilized this dataset to fine-tune the GPT-3.5 model\footnote{\url{https://openai.com/blog/gpt-3-5-turbo-fine-tuning-and-api-updates}}. Subsequently, the trained GPT-3.5 model was employed to automatically annotate an additional 1,000 unlabeled data samples.

To fix our results, we selected the GPT-3.5-Turbo-0613 version for the fine-tuning process. The fine-tuning incorporated a blend of In-context Learning (ICL) and Instruction Learning (IL) methodologies for the input format. For ICL, a moderately lengthy sample was randomly chosen as a sequence. For IL, an instructive prompt was strategically placed between the sample and the sentence set for categorization. This prompt guides the model to classify and extract subsequent text based on the preceding sample. Further details on ICL and IL will be elaborated upon in the upcoming "USA Model Training" section.

Upon completing this process, the model successfully auto-generated 1,000 accurately labeled datasets. We subsequently conducted a manual review of these 1,000 samples, rectifying any incorrect labels. This streamlined approach significantly reduced the manual labor and time investment.

\subsection{SCPOS dataset of Rule-based Matching}

In this section, we outline the method employed to use a Japanese word polarity dictionary\footnote{\url{https://www.cl.ecei.tohoku.ac.jp/Open_Resources-Japanese_Sentiment_Polarity_Dictionary.html}} for matching corresponding words within a sentence. We subsequently constructed three SCPOS datasets, each containing different parts-of-speech (POS) classifications.

Initially, we utilized two distinct word polarity dictionaries. The first dictionary comprises 13,264 prevalent Japanese nouns, categorized into three polarity classes: positive, neutral, and negative. In contrast, the second dictionary has 5,280 frequently used Japanese verbs and adjectives, classified into two polarity groups: positive and negative. Figure \ref{figure3} illustrates the procedure of using these dictionaries for rule-based matching against the training set of the MARC-ja dataset. It's worth noting that if identical words are repeated within the text, each occurrence is identified and listed sequentially. Following this process, we derived two SCPOS datasets labeled according to their respective POS: Noun (N) and Verb \& Adjective (VA). Furthermore, by amalgamating both dictionaries, we crafted a comprehensive polarity thesaurus covering three POS: Noun, Verb, and Adjective. This merged thesaurus was then matched against the MARC-ja dataset, resulting in the creation of the SCPOS dataset encompassing Noun, Verb, and Adjective (NVA).

Consequently, all the derived datasets not only retain the original text sentiment classification label but also incorporate word polarity classifications.

\begin{figure*}[!t]
\centering
\includegraphics[width=438 pt]{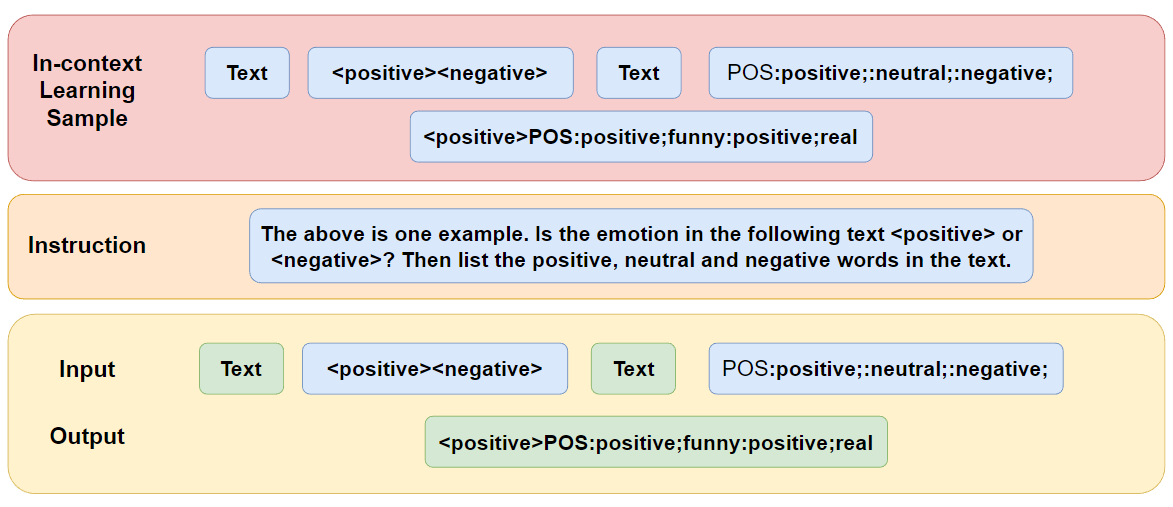}
\caption{Example of In-context Learning and Instruction Learning of input format.}
\label{figure5}
\end{figure*}

\section{Universal Sentiment Analysis Model}\label{USA model}
In this chapter, we present the Universal Sentiment Analysis Model (USA).  The discussion is organized into three subsections to elucidate the process systematically.  First, we address the preparation of the training corpus, followed by the conversion of input and output formats.  We conclude with the implementation of ICL and IL techniques, which are essential for training an LLM to achieve robust performance in sentiment analysis tasks.

\subsection{Constrcution of Train Corpus}

To fine-tune the LLM, we initially prepared a specialized corpus. As depicted in Figure \ref{figure3}, the green block represents the corpus constructed for this purpose. We extracted three sub-datasets, each containing 500 samples, from the MARC-ja dataset's test set. A consistent rule-based method was employed for both matching and generation. When combined with the 1,000 samples previously generated during the fine-tuning of the GPT-3.5 model, we amassed a composite dataset of 2,500 samples, encompassing four distinct subtasks.

Our primary objective was to maximize the model's learning from the SRW sub-dataset. This emphasis on the SRW dataset is because it solely concentrates on words that align with the overall sentiment polarity of the text, making it an optimal choice for teaching the model about words that significantly influence the text's sentiment. Additionally, the SRW dataset boasts exceptionally high labeling quality. Consequently, we doubled the weight of the SRW sub-dataset in the training corpus compared to the other datasets.

Previous research on large model fine-tuning suggests that only a minimal amount of high-quality data is required to effectively fine-tune an LLM\cite{zhou2023lima}. In alignment with this understanding, we used a mere 2,500 samples to train the LLM. Utilizing fewer samples for fine-tuning also ensures better preservation of knowledge acquired by the LLM during its pre-training phase.

\subsection{Format of Input and Output}

After preprocessing the training corpus, it's essential to address the input and output data formats. We bifurcate the role of the Format Converter(FC) into two primary functions. First, the FC standardizes the input and output formats across all tasks, ensuring that both the text sentiment polarity classification and the word sentiment polarity classification tasks utilize a consistent format. Secondly, the FC serves the function of providing a prompt. The concept of a prompt was introduced to guide the model in generating desired labels with higher accuracy in few-shot scenarios. In our research, we consistently structure both input and output using fixed symbols and words. This strategy aims to employ these symbols and words as cues, guiding the model to produce tokens corresponding to the desired labels following these cues.

According to the two roles proposed earlier, we refer to the structure of FC in SLG Framework\cite{10.1007/978-3-031-35320-8_18}. Figure \ref{figure4} illustrates the transformation of the input and output format, using the original dataset. The input comprises the text and its corresponding overall sentiment polarity categorization labels. Additionally, the output provides polarity labels for individual words and their respective word spans. For simplicity in subsequent discussions, we will refer to these as 'PW pairs'. These PW pairs maintain a one-to-one correspondence between the labels and word pairs.

In the lower segment of Figure \ref{figure4}, the FC-converted input and output formats are depicted. The blue block represents a fixed sequence of tags and words in the input. Initially, the text for categorization is input, followed by the addition of two text polarity classification labels. This text is then re-input, and the cue word "POS" is concatenated with the next three word polarity classification labels. Together, these elements form a comprehensive input sequence for the model.

In the output section, text sentiment polarity classification labels are enclosed by pink background symbols. The trailing "POS" serves as an indicator for the model to begin producing PW pairs. The characters ":", set against a dark blue background, and ";" signify the start and finish of word sentiment polarity labeling, respectively. With this, the transition of the input and output format is effectively completed.

\subsection{In-context Learning and Instruction Learning}
In this section, we elucidate the process by which the 'transformed format' dataset is packaged by ICL and IL, and subsequently fed into the model. Figure \ref{figure5} illustrates the sequence: the ICL sample is at the top, followed by the IL instruction problem, and lastly, the input in its transformed format. Blue blocks denote sequences consistent across all input samples, while green blocks indicate segments replaced based on specific input samples.

Focusing first on the ICL component: a medium-length text was randomly selected from the MARC-ja test set, ensuring avoidance of the 1500 training phrases previously extracted. We intentionally did not choose a lengthy text encompassing a wide array of POS samples. The chosen samples were labeled following the conventions of the prior four datasets. This labeling was then formatted into the final ICL sample using the FC format.

Turning to the instructional segment, it's divided into three distinct sentences. First, in the first sentence we hint that the sequence in front of the model is an example. The second sentence asks whether the overall sentiment polarity classification of the text given next is positive or negative. Here again, marks are placed on both sides of the positive and negative words. Finally, the third sentence asks the model to list the words in the text that are related to the polarity of the sentiment. This instruction requires modifications based on the specifics of four datasets. As an instance, the version portrayed in Figure \ref{figure5} corresponds to the SRW sub-dataset. For the NVA dataset, "word" should be replaced with "noun, verb, and adjective".

In summation, the combination of the ICL sample problem and Instruction, along with the Input, establishes the sequence for model input, leading to the output generation. Employing IL serves as a prompt, aligning the downstream task with the instruction fine-tuning phase in LLM pre-training. Simultaneously, the integration of ICL potentially extends instructional length, thereby refining the generative outcomes of LLMs.

\section{Evaluation}\label{evaluation}
In this chapter, we delineate the experimental framework and define the evaluation metrics. Subsequently, we assess the chosen models employing the SCPOS dataset.

\subsection{Experiment Set}

To begin, due to constraints in time and resources, we opted for a sample size of 1,000, randomly selected from all SCPOS sub-datasets. Each sample was tested three times, and the average value was taken as the final result. For model selection that the LLaMA2-7B model served as the base model for the USA model. Under the USA-7B model umbrella, we trained two distinct versions:
1.USA-7B(ICL+IL) using the combined ICL and IL data format .
2.USA-7B(IL) solely utilizing the IL data format.
For comparative purposes, we incorporated the T5-base-Japanese model, which operates on the SLG framework. Additionally, the original gpt-3.5-turbo model was chosen as a benchmark for comparison.
It's imperative to note that our selection of comparative models wasn't arbitrary. We rigorously tested a multitude of LLMs trained on Japanese corpora, as well as the LLaMA2-70B model. These models were subjected to both 1-shot ICL and IL evaluations. Regrettably, none of the models produced satisfactory results. They frequently generated text echoing the input or text irrelevant to the task. Consequently, our final comparison was limited to the T5 and gpt-3.5-turbo models.

The T5 model lacks the capability for few-shot learning in tasks with long sequences. Therefore, for training the T5 model, we utilized 1,000 randomly selected samples, while another set of 1,000 samples served as the test set. In the SRW task, we used the GPT-3.5-generated data for training and a manually labeled dataset for testing.

Upon evaluating the GPT-3.5 model, we observed that simultaneous use of ICL and IL during testing essentially prevents the generation of the desired text. Consequently, for testing GPT-3.5, only ICL was employed to input the dataset into the model.

\begin{table*}[!t]
\centering
\renewcommand{\arraystretch}{1.3}
\begin{tabular}{lcccccc}
\hline
  &  & \textbf{SRW} &  &  & \textbf{NVA} &     \\
\hline
 \textbf{Accuracy} & \textbf{$ACC_{SC}$} & \textbf{$ACC_{POS}$} & \textbf{$ACC_{SCPOS}$} &\textbf{$ACC_{SC}$} & \textbf{$ACC_{POS}$} & \textbf{$ACC_{SCPOS}$}   \\
\hline
\textbf{T5-base(SLG)} & 88.21 & 55.57 & 17.28 & 87.30 & 26.22 & 1.60      \\
\textbf{USA-7B(ICL+IL)} & \textbf{89.60} & \textbf{56.32} & \textbf{18.10} & \textbf{90.20} & \textbf{60.09} & \textbf{3.97}     \\
\textbf{USA-7B(IL)} & 88.60 & 53.24 & 17.50 & 88.43 & 55.28 & 3.60     \\
\textbf{GPT-3.5}& 53.6 & 14.99 & 1.60 & 73.20 & 10.34 &  0.13   \\
\hline
&  & \textbf{Nous} &  &  & \textbf{Verb \& Adjective} & \\
\hline
&\textbf{$ACC_{SC}$} & \textbf{$ACC_{POS}$} & \textbf{$ACC_{SCPOS}$} &\textbf{$ACC_{SC}$} & \textbf{$ACC_{POS}$} & \textbf{$ACC_{SCPOS}$} \\
\hline
\textbf{T5-base(SLG)} & 89.50 & 27.62 & 3.00 & 83.00 & \textbf{73.84} & \textbf{52.47}   \\
\textbf{USA-7B(ICL+IL)} & \textbf{91.50} & \textbf{62.41} & \textbf{6.83} & \textbf{92.17} & 64.94 & 50.90    \\
\textbf{USA-7B(IL)} & 90.80 & 57.74 & 4.73 & 89.33 & 69.83 & 52.43    \\
\textbf{GPT-3.5}& 73.83 & 10.44 & 0.23 & 78.83 & 15.45 & 9.87     \\
\hline

\hline
\end{tabular}
\caption{\label{table2SCPOSresult}
Results for the four SCPOS subdatasets on the four models. T5-base was fine-tuned. USB-7B(ICL+IL) was assessed with a 1-shot approach. USB-7B(IL) utilized a 0-shot approach. GPT-3.5 was tested using the 1-shot ICL method.}
\end{table*}

\subsection{Evaluation Metrics}

In the selection of evaluation metrics, because of the uncertainty of the results generated by the generative model. It leads to the length of generated sequences is not fixed like sequence labeling models. So it is difficult to use metrics such as F1 to evaluate the task. Especially in this task, some texts contain as many as 20-30 PW pairs. Here accuracy is used as the evaluation metrics. Three different accuracies are categorized according to the type of task.

The first one is the text sentiment polarity classification accuracy(i.e. $ACC_{SC}$). the characters in the text classification part of the generated sequence and the actual text are intercepted and compared, and all of them are considered to be correct if they are equal (e. g. Generated: <positive> = Actual: <positive>). 

Next is the calculation of the accuracy of the remaining part of the PW pair. All the PW pairs of the generated and actual sequences are partitioned according to the ":" and ";" notations. Let the set $P$ contain $n$ PW pairs ($X_{1}$, $X_{2}$, ...., $X_{n}$). Two sets can be obtained, PW pair $P_{generated}$ for the generated sequence and PW pair $P_{actual}$ for the actual sequence, and then the sets$P_{generated}$ and $P_{actual}$ are matched. The number of matched pairs is divided by the total number of PW pairs of the actual sequence to get the $ACC_{pos}$. 

Finally the SCPOS accuracy is calculated. When both the $ACC_{text}$ and the $ACC_{pos}$ of the same sample are equal to 1, it is computed as a correct sample. The number of all correct samples is divided by the total number of samples to get the $ACC_{SCPOS}$.

\subsection{Results}

As illustrated in Table \ref{table2SCPOSresult}, the four models were tested on the four sub-datasets. It is evident that USB-7B(ICL+IL) outperforms the others by achieving the highest accuracy in 10 out of 12 evaluations across these datasets. This underscores its robust capability in the SCPOS task. Notably, both USB-7B(ICL+IL) and USB-7B(IL) models significantly surpass the accuracy of the GPT-3.5 model.

Historically, in the extraction and classification of constructed data, Large Language Models (LLMs) have been overshadowed by smaller Pretrained Language Models (PLMs), as evidenced by the results of GPT-3.5. Nevertheless, after fine-tuning with the SCPOS mixed training corpus, the USA models demonstrate an improved ability to outshine their counterparts. Leveraging the extensive pre-training corpus and parameter set of LLMs, they can surpass PLMs that are exclusively fine-tuned with the entirety of sub-tasks in 0-shot scenarios.  This also indicates that the finely-tuned USB-7B model, even with fewer samples, can exceed the performance of the full-volume fine-tuned T5-base.

In the VA sub-datasets, the USA model did not surpass the performance of the fine-tuned T5 model. The reason for this disparity lies in the fact that the average input and output lengths in the VA dataset are considerably shorter than in the other three datasets. Consequently, the fine-tuned T5 model demonstrates superior results compared to the 0/1-shot USA model when handling short sequences.

Furthermore, the performance of the USA-7B model augmented with ICL is superior to its counterpart using IL exclusively. This suggests that the combined use of ICL and IL enhances the model's comprehension of each sub-dataset. Moreover, distinct ICL and IL were employed for each of the four sub-datasets. This approach enhances the model's ability to differentiate and comprehend the various tasks.

\section{Conclusion}

In this paper, we introduces a novel sentiment categorization task termed the SCPOS task and details the creation of four distinct sub-datasets. Utilizing the SCPOS task, we trained the USA-7B model, achieving exemplary performance in 0/few-shot sentiment classification scenarios. We anticipate that the proposed SCPOS dataset and the USA-7B model will pave the way for fresh research avenues and areas of focus in sentiment analysis.

In future research, we plan to evaluate additional LLMs using the SCPOS dataset. Our objectives are to ascertain the presence of MRE in both the text sentiment classification and the POS sentiment polarity classification tasks. Furthermore, we aim to determine whether LLMs leverage MRE to enhance the performance of these tasks.

\bibliography{anthology,custom}

\begin{thebibliography}{22}
\expandafter\ifx\csname natexlab\endcsname\relax\def\natexlab#1{#1}\fi

\bibitem[{Agarwal et~al.(2011)Agarwal, Xie, Vovsha, Rambow, and Passonneau}]{agarwal2011sentiment}
Apoorv Agarwal, Boyi Xie, Ilia Vovsha, Owen Rambow, and Rebecca~J Passonneau. 2011.
\newblock Sentiment analysis of twitter data.
\newblock In \emph{Proceedings of the workshop on language in social media (LSM 2011)}, pages 30--38.

\bibitem[{Bai(2011)}]{BAI2011732}
Xue Bai. 2011.
\newblock \href {https://doi.org/https://doi.org/10.1016/j.dss.2010.08.024} {Predicting consumer sentiments from online text}.
\newblock \emph{Decision Support Systems}, 50(4):732--742.
\newblock Enterprise Risk and Security Management: Data, Text and Web Mining.

\bibitem[{Brown et~al.(2020)Brown, Mann, Ryder, Subbiah, Kaplan, Dhariwal, Neelakantan, Shyam, Sastry, Askell et~al.}]{brown2020language}
Tom Brown, Benjamin Mann, Nick Ryder, Melanie Subbiah, Jared~D Kaplan, Prafulla Dhariwal, Arvind Neelakantan, Pranav Shyam, Girish Sastry, Amanda Askell, et~al. 2020.
\newblock Language models are few-shot learners.
\newblock \emph{Advances in neural information processing systems}, 33:1877--1901.

\bibitem[{Devlin et~al.(2018)Devlin, Chang, Lee, and Toutanova}]{devlin2018bert}
Jacob Devlin, Ming-Wei Chang, Kenton Lee, and Kristina Toutanova. 2018.
\newblock Bert: Pre-training of deep bidirectional transformers for language understanding.
\newblock \emph{arXiv preprint arXiv:1810.04805}.

\bibitem[{Gan et~al.(2023)Gan, Zhang, and Mori}]{10.1007/978-3-031-35320-8_18}
Chengguang Gan, Qinghao Zhang, and Tatsunori Mori. 2023.
\newblock Sentence-to-label generation framework for multi-task learning of japanese sentence classification and named entity recognition.
\newblock In \emph{Natural Language Processing and Information Systems}, pages 257--270, Cham. Springer Nature Switzerland.

\bibitem[{Go et~al.(2009)Go, Bhayani, and Huang}]{go2009twitter}
Alec Go, Richa Bhayani, and Lei Huang. 2009.
\newblock Twitter sentiment classification using distant supervision.
\newblock \emph{CS224N project report, Stanford}, 1(12):2009.

\bibitem[{Higashiyama et~al.(2008)Higashiyama, Inui, and Matsumoto}]{Adjectives}
Masahiko Higashiyama, Kentaro Inui, and Yuji Matsumoto. 2008.
\newblock Learning sentiment of nouns from selectional preferences of verbs and adjectives.
\newblock \emph{Proceedings of the 14th Annual Meeting of the Association for Natural Language Processing}, 4:584--587.

\bibitem[{Kobayashi et~al.(2005)Kobayashi, Inui, Matsumoto, and Tateishi}]{2005203}
Nozomi Kobayashi, Kentaro Inui, Yuji Matsumoto, and Kenji Tateishi. 2005.
\newblock \href {https://doi.org/10.5715/jnlp.12.3_203} {Collecting evaluative expressions for opinion extraction}.
\newblock \emph{Journal of Natural Language Processing}, 12(3):203--222.

\bibitem[{Kurihara et~al.(2022)Kurihara, Kawahara, and Shibata}]{kurihara2022jglue}
Kentaro Kurihara, Daisuke Kawahara, and Tomohide Shibata. 2022.
\newblock Jglue: Japanese general language understanding evaluation.
\newblock In \emph{Proceedings of the Thirteenth Language Resources and Evaluation Conference}, pages 2957--2966.

\bibitem[{Li et~al.(2023)Li, Lin, Wu, Liu, Tang, Zhao, and Li}]{li2023unisa}
Zaijing Li, Ting-En Lin, Yuchuan Wu, Meng Liu, Fengxiao Tang, Ming Zhao, and Yongbin Li. 2023.
\newblock \href {http://arxiv.org/abs/2309.01339} {Unisa: Unified generative framework for sentiment analysis}.

\bibitem[{Medhat et~al.(2014)Medhat, Hassan, and Korashy}]{medhat2014sentiment}
Walaa Medhat, Ahmed Hassan, and Hoda Korashy. 2014.
\newblock Sentiment analysis algorithms and applications: A survey.
\newblock \emph{Ain Shams engineering journal}, 5(4):1093--1113.

\bibitem[{OpenAI(2023)}]{openai2023gpt4}
OpenAI. 2023.
\newblock \href {http://arxiv.org/abs/2303.08774} {Gpt-4 technical report}.

\bibitem[{Ouyang et~al.(2022)Ouyang, Wu, Jiang, Almeida, Wainwright, Mishkin, Zhang, Agarwal, Slama, Ray, Schulman, Hilton, Kelton, Miller, Simens, Askell, Welinder, Christiano, Leike, and Lowe}]{ouyang2022training}
Long Ouyang, Jeff Wu, Xu~Jiang, Diogo Almeida, Carroll~L. Wainwright, Pamela Mishkin, Chong Zhang, Sandhini Agarwal, Katarina Slama, Alex Ray, John Schulman, Jacob Hilton, Fraser Kelton, Luke Miller, Maddie Simens, Amanda Askell, Peter Welinder, Paul Christiano, Jan Leike, and Ryan Lowe. 2022.
\newblock \href {http://arxiv.org/abs/2203.02155} {Training language models to follow instructions with human feedback}.

\bibitem[{Pang et~al.(2002)Pang, Lee, and Vaithyanathan}]{pang2002thumbs}
Bo~Pang, Lillian Lee, and Shivakumar Vaithyanathan. 2002.
\newblock Thumbs up? sentiment classification using machine learning techniques.
\newblock In \emph{Proceedings of the 2002 Conference on Empirical Methods in Natural Language Processing (EMNLP 2002)}, pages 79--86.

\bibitem[{Raffel et~al.(2020)Raffel, Shazeer, Roberts, Lee, Narang, Matena, Zhou, Li, and Liu}]{raffel2020exploring}
Colin Raffel, Noam Shazeer, Adam Roberts, Katherine Lee, Sharan Narang, Michael Matena, Yanqi Zhou, Wei Li, and Peter~J Liu. 2020.
\newblock Exploring the limits of transfer learning with a unified text-to-text transformer.
\newblock \emph{The Journal of Machine Learning Research}, 21(1):5485--5551.

\bibitem[{Robaldo and {Di Caro}(2013)}]{ROBALDO2013454}
Livio Robaldo and Luigi {Di Caro}. 2013.
\newblock \href {https://doi.org/https://doi.org/10.1016/j.csi.2012.10.004} {Opinionmining-ml}.
\newblock \emph{Computer Standards \& Interfaces}, 35(5):454--469.

\bibitem[{Vaswani et~al.(2017)Vaswani, Shazeer, Parmar, Uszkoreit, Jones, Gomez, Kaiser, and Polosukhin}]{vaswani2017attention}
Ashish Vaswani, Noam Shazeer, Niki Parmar, Jakob Uszkoreit, Llion Jones, Aidan~N Gomez, {\L}ukasz Kaiser, and Illia Polosukhin. 2017.
\newblock Attention is all you need.
\newblock \emph{Advances in neural information processing systems}, 30.

\bibitem[{Wu and Tan(2011)}]{wu2011two}
Qiong Wu and Songbo Tan. 2011.
\newblock A two-stage framework for cross-domain sentiment classification.
\newblock \emph{Expert Systems with Applications}, 38(11):14269--14275.

\bibitem[{Yu et~al.(2013)Yu, Duan, and Cao}]{yu2013impact}
Yang Yu, Wenjing Duan, and Qing Cao. 2013.
\newblock The impact of social and conventional media on firm equity value: A sentiment analysis approach.
\newblock \emph{Decision support systems}, 55(4):919--926.

\bibitem[{Zhang et~al.(2018)Zhang, Wang, and Liu}]{zhang2018deep}
Lei Zhang, Shuai Wang, and Bing Liu. 2018.
\newblock Deep learning for sentiment analysis: A survey.
\newblock \emph{Wiley Interdisciplinary Reviews: Data Mining and Knowledge Discovery}, 8(4):e1253.

\bibitem[{Zhou et~al.(2023{\natexlab{a}})Zhou, Liu, Xu, Iyer, Sun, Mao, Ma, Efrat, Yu, Yu, Zhang, Ghosh, Lewis, Zettlemoyer, and Levy}]{zhou2023lima}
Chunting Zhou, Pengfei Liu, Puxin Xu, Srini Iyer, Jiao Sun, Yuning Mao, Xuezhe Ma, Avia Efrat, Ping Yu, Lili Yu, Susan Zhang, Gargi Ghosh, Mike Lewis, Luke Zettlemoyer, and Omer Levy. 2023{\natexlab{a}}.
\newblock \href {http://arxiv.org/abs/2305.11206} {Lima: Less is more for alignment}.

\bibitem[{Zhou et~al.(2023{\natexlab{b}})Zhou, Zhang, Gu, Chen, and Poon}]{zhou2023universalner}
Wenxuan Zhou, Sheng Zhang, Yu~Gu, Muhao Chen, and Hoifung Poon. 2023{\natexlab{b}}.
\newblock Universalner: Targeted distillation from large language models for open named entity recognition.
\newblock \emph{arXiv preprint arXiv:2308.03279}.

\end{thebibliography}
\bibliographystyle{acl_natbib}

\appendix

\section{Analysis of Mutual Reinforcement Effect}\label{MRE}

\begin{table*}[!ht]
\centering
\renewcommand{\arraystretch}{1.3}
\begin{tabular}{lcccccc}
\hline
  &  & \textbf{SRW} &  &  & \textbf{NVA} &     \\
\hline
 \textbf{Accuracy} & \textbf{$ACC_{SC}$} & \textbf{$ACC_{POS}$} & \textbf{$ACC_{SCPOS}$} &\textbf{$ACC_{SC}$} & \textbf{$ACC_{POS}$} & \textbf{$ACC_{SCPOS}$}   \\
\hline
\textbf{SCPOS} & 88.21 & \textbf{55.57} & \textbf{17.28} & 87.30 & \textbf{26.22} & \textbf{1.60}      \\
\textbf{SC Only} & \textbf{92.91} & - & - & \textbf{91.40} & - & -     \\
\textbf{POS Only} & - & 30.69 & - & - & 24.65 & -     \\
\hline
&  & \textbf{Nous} &  &  & \textbf{Verb \& Adjective} & \\
\hline
&\textbf{$ACC_{SC}$} & \textbf{$ACC_{POS}$} & \textbf{$ACC_{SCPOS}$} &\textbf{$ACC_{SC}$} & \textbf{$ACC_{POS}$} & \textbf{$ACC_{SCPOS}$} \\
\hline
\textbf{SCPOS} & 89.50 & \textbf{27.62} & \textbf{3.00} & 83.00 & \textbf{73.84} & \textbf{52.47}   \\
\textbf{SC Only} & \textbf{91.40} & - & - & \textbf{91.40} & - & -    \\
\textbf{POS Only} & - & 24.05 & - & - & 52.04 & -    \\
\hline

\hline
\end{tabular}
\caption{\label{table3MRE}
The results of the SC and POS tasks were tested separately using the T5-base model.}
\end{table*}

To determine if MRE is prevalent in both the SC and POS sentiment polarity classification tasks, we carried out distinct evaluations for each task. To ensure consistency and control for other variables, we employed the T5-base model across all four datasets. As evidenced in Table 3, when the SC task is executed independently, the performance across all datasets surpasses that of the SCPOS task. This is likely because the SC task necessitates generating only a brief text categorization label, leading to enhanced accuracy. In contrast, when the POS task is conducted independently, there's a marked decline in accuracy across all datasets. This further substantiates the presence of MRE in the sentiment polarity classification task.

\section{Details of Experiment}
To train the USA-7B model, we employed four NVIDIA A800 GPUs, each with 80GB, running for approximately 20 hours. We trained two models for durations of 2 and 3 epochs, respectively, using a learning rate of 1e-5. Collectively, they consumed between 160-200GB of GPU memory.

For the dataset auto-annotation task, we utilized the fine-tuned GPT-3.5 model. Our dataset comprised 1,000 data points, amounting to 4.5 million tokens, and the associated cost was approximately \$57. We conducted 3 epochs of training for this task. Unless specified, we maintained the default settings provided by OpenAI for all other training parameters. In subsequent training and testing experiments involving the T5 model, we used an RTX 3090 GPU with 24GB.
Regarding the generation parameter settings for the models. The USA-7B model was set with the following parameters:
\begin{itemize}
    \item max new tokens: 2000
    \item repetition penalty: 1.3
    \item temperature: 1.0
    \item top\_p: 0.7
    \item top\_k: 40
\end{itemize}
For both the T5 and GPT-3.5 models, we only adjusted the "max new tokens" setting to 400. It's worth noting that the average output length of the SCPOS dataset, which boasts the longest average, is 144 tokens. All other parameters remained at their default values.

\end{document}